**Daniel Hopp**
Associate Statistician
Division on Globalisation and Development Strategies, UNCTAD

daniel.hopp@unctad.org

# Benchmarking Econometric and Machine Learning Methodologies in Nowcasting

## Abstract

Nowcasting can play a key role in giving policymakers timelier insight to data published with a significant time lag, such as final GDP figures. Currently, there are a plethora of methodologies and approaches for practitioners to choose from. However, there lacks a comprehensive comparison of these disparate approaches in terms of predictive performance and characteristics. This paper addresses that deficiency by examining the performance of 12 different methodologies in nowcasting US quarterly GDP growth, including all the methods most commonly employed in nowcasting, as well as some of the most popular traditional machine learning approaches. Performance was assessed on three different tumultuous periods in US economic history: the early 1980s recession, the 2008 financial crisis, and the COVID crisis. The two best performing methodologies in the analysis were long short-term memory artificial neural networks (LSTM) and Bayesian vector autoregression (BVAR). To facilitate further application and testing of each of the examined methodologies, an open-source repository containing boilerplate code that can be applied to different datasets is published alongside the paper, available at: github.com/dhopp1/nowcasting_benchmark.

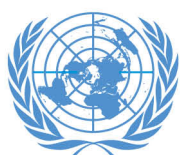

# Table of Contents

## Acknowledgements

I would like to thank Anu Peltola for her valuable comments and feedback.

# 1. Introduction

Gross domestic product's (GDP) importance in quantifying the size and performance of the economy cannot be understated. It is the go-to metric for government officials, policymakers, and indeed even the general public for insight to economic health, and by extension a myriad of related measures, such as social well-being (Dynan et al., 2018; IMF, 2020; Kapoor and Debroy, 2019). However, as much as we depend on this figure, the reality is that it is often published with a significant lag. This lag depends on the country, but in the United States, advanced estimates for an elapsed quarter are not released for at least one month afterwards, with final estimates not appearing until three months afterwards (Federal Reserve Bank of San Francisco, 2005). This delay is related to the complexity of calculating GDP relative to other indicators, with its myriad of sources and adjustments. Policymakers can turn to other, faster-publishing indicators for a quicker look into the state of the economy, such as prices or industrial production, but GDP's comprehensive nature, that which simultaneously delays its publication, is precisely what makes it desirable as an indicator.

Given these characteristics, the utility of applying nowcasting to the case of GDP becomes clear. Nowcasting "is the estimation of the current, or near to it either forwards or backwards in time, state of a target variable using information that is available in a timelier manner" (Hopp, 2021a). In essence, series that are available in a timelier manner than GDP can be used to estimate a model using historical data, when data for both GDP and these independent series are available. This model can then be used to obtain estimates for GDP well before advanced estimates are available, even while the quarter for prediction is ongoing. GDP's publication lag and salience as an indicator have rendered it perhaps the most common target variable for nowcasting applications. This makes it an ideal choice for a benchmark analysis comparing different nowcasting approaches.

The need for such an analysis stems from the existence of several disparate methodologies specifically employed for nowcasting, in addition to other commonly used econometric and machine learning techniques. A new practitioner, or an experienced practitioner looking to improve their models or expand to different datasets, is currently hard-pressed to find a starting point or an overview of nowcasting approaches when making their modelling decisions. The need to refer to several papers to get a sense of different methodologies' performance and characteristics, which may be applied to different datasets, can obfuscate conclusions. It is thus the goal of this paper to consolidate results of the most common statistical, econometric, and machine learning methodologies in nowcasting using perhaps the closest thing to a benchmark dataset available in the field; nowcasting quarterly US GDP growth using explanatory variables from the Federal Reserve of Economic Data (FRED) as specified in Bok et al. (2018). This dataset has the additional benefit of a long publication history, dating back to 1947. This allows performance to be tested on three separate periods in American history with exceptional economic circumstances: the early 1980s recession, the 2008 financial crisis, and the COVID crisis. These three periods are used to test the performance of 12 different methodologies in nowcasting GDP growth. Detailed information on each is provided in section 2.2.

In alphabetical order, they are:

- autoregressive-moving-average (ARMA)
- Bayesian mixed-frequency vector autoregression (Bayesian VAR)
- decision trees
- dynamic factor models (DFM)
- gradient boosted trees
- long short-term memory artificial neural networks (LSTM)
- mixed data sampling regression (MIDAS)
- mixed-frequency vector autoregression (MF-VAR)
- multilayer perceptron feedforward artificial neural networks (MLP)
- ordinary least squares regression (OLS)
- random forest
- ridge regression

The primary goal of this paper is not only to shed light on these methods' relative performance and characteristics in nowcasting, but also to enable practitioners to take these findings and apply them to their own data. Consequently, an accompanying open-source repository has been created where boilerplate code in Python or R for each methodology can be found and easily adapted to different datasets (Hopp, 2022). With this tool, the barrier to trying out nowcasting on different applications may be lowered, as well as the barrier to trying out multiple methodologies to validate results and increase the chances of obtaining a well-performing model.

The rest of the paper will proceed in the following manner: section two will provide further background on nowcasting and the methodologies employed; section three will detail the data used; section four will explain the modelling approach and how results were obtained; section five will display and discuss results; section six will conclude.

# 2. Background

## 2.1 Nowcasting

The term nowcasting was first coined and applied in the meteorological domain in the early 1980s to describe weather forecasting of the near future with information on current meteorological conditions (WMO, 2017). It did not begin appearing in economic literature until the mid-2000s, where the term became popularized after the publication of Giannone et al. (2005). The concept of obtaining real-time, data-based estimates of the macroeconomic situation predates 2005, however, with Mariano and Murasawa (2003) developing a coincident business cycle index based on monthly and quarterly series. This application was already very similar to what is considered economic nowcasting today, but did not directly nowcast GDP, rather equating its synthesized business index to "the smoothed estimate of latent monthly real GDP" (Mariano and Murasawa, 2003). Post-2005, a wealth of papers began to be published examining nowcasting different combinations of indicators, most commonly GDP, and geographies. Examples include Portuguese GDP (Morgado et al., 2007), European GDP (Giannone et al., 2009), global trade (Cantú, 2018; Guichard and Rusticelli, 2011), and German GDP (Marcellino and Schumacher, 2010).

A further differentiating axis in the nowcasting literature, and that of primary concern in this paper, is the methodological approach employed. The most commonly-used approach in nowcasting is perhaps the DFM, which is employed in a wealth of papers, including in Giannone et al. (2005). Other examples include Antolin-Diaz et al. (2020), Cantú (2018), and Guichard and Rusticelli (2011), among many others. Other commonly employed approaches include MIDAS (Kuzin et al., 2009; Marcellino and Schumacher, 2010), MF-VAR (Kuzin et al., 2009), Bayesian VAR (Cimadomo et al., 2020), LSTMs (Hopp, 2021a, 2021b), and many more. Each of these methodologies has characteristics which make them suitable for use in the nowcasting context, which comes with its own particular data challenges, discussed below. But Richardson et al. (2021) additionally examined using common machine learning algorithms in predicting New Zealand GDP growth.

It is drawing upon this literature that the 12 methodologies examined in this analysis were chosen: Bayesian VAR, DFM, LSTM, MF-VAR, MIDAS, and MLP were all included as they appear frequently in the nowcasting literature; ARMA was included as a baseline model; OLS and ridge regression were included as perhaps the most popular regression technique in the case of the former and as an augmentation of OLS which could render it more suitable for nowcasting in the case of the latter; decision trees, gradient boosted trees, and random forest were included as three popular machine learning techniques (Sarker, 2021).

As mentioned earlier, nowcasting comes with its own set of data challenges, which each methodology needed to be able to handle. Details of this process for each methodology are outlined in the next section. The first challenge is mixed-frequency data, where all variables in the model are not recorded in the same frequency. In this analysis, for example, a mixture of quarterly and monthly variables was used to estimate a quarterly variable, GDP growth. The second is "ragged-edges", or differences in missing variables at the end of series due to different publication schedules for each. The model needs some way to be able to handle partially complete data at the ends of time series. The third is the "curse of dimensionality", where there may be relatively more input variables to a model compared with training observations (Buono et al., 2017). This can lead to estimation and other errors in some methodologies, such as causing multicollinearity in OLS. If nowcasting is ever to leverage the power of big data and not be restricted to a handful of explanatory variables, this last challenge will be of particular importance.

## 2.2 Methodologies

The following sections will provide background information as well as references for further reading for each methodology. Due to the quantity of methodologies included in the analysis, it is not possible to include a comprehensive quantitative explanation of each. For those interested, in-depth explanations of this nature are available via the references. The particular programming implementations utilized for each will also be discussed. Methodologies are presented in alphabetical order.

### 2.2a ARMA

ARMA models are the simplest nowcasting approach examined in this paper, modelling a time series in terms of two main elements: an autoregressive (AR) component, where future values in a series are a function of its own $p$ prior values, and a moving-average (MA) component, where the error terms of the series are a function of $q$ prior error terms. For more information on the use of ARMA models for modelling stationary time series, see Mills (2019). ARMA is a univariate approach for modelling a time series, meaning that, unlike the other 11 methodologies, the ARMA model included only GDP growth's own history as an input variable. This parsimonious nature makes the model an attractive and commonly-used benchmark in nowcasting applications, such as in Cimadomo et al. (2020) or Ministry of Transport, New Zealand Government (2016).

For this analysis, the *auto.arima* function of the *pmdarima* (Smith, 2021) Python library was used to determine the $p$ and $q$ lag orders of the ARMA model on the training set. See section three for more information on the meaning of "training set". The *auto.arima* function determines the lag orders by fitting models with different lag permutations and recording their Akaike Information Criteria (AIC), then selecting the orders which minimize this value. See Liew (2004) for more information on AIC as well as lag selection in ARMA models in general. The *ARIMA* function of *pmdarima* was then used to fit and generate final predictions.

### 2.2b Bayesian mixed-frequency vector autoregression

Standard vector autoregression (VAR) is similar to the univariate AR model discussed previously, but generalized to consider multiple time series. Whereas in the univariate case, a variable's value is a function of its $p$ prior values, in the multivariate case, a set of variables' prior values are a function of the set's prior values. Essentially, a vector is considered in the modelling rather than a scalar. For more information on VAR models, see (Stock and Watson, 2001).

In contrast to standard VARs, where model parameters are estimated and taken as fixed values, Bayesian VARs consider the parameters as random variables with an assigned prior probability. This approach helps to mitigate over-parameterization, the third data issue discussed in section 2.1. Standard VARs struggle with this issue due to the high number of parameters required to estimate them, usually restricting their use to applications with less than 10 input variables (Bańbura et al., 2010). The introduction of Bayesian shrinkage has been shown to increase forecast accuracy in VAR models with as little as six input variables, many fewer than may be found in a typical nowcasting model. For more information on the concepts of Bayesian shrinkage and Bayesian statistics as applied to regression problems, see De Mol et al. (2008). Bayesian VAR's power in modelling complex dynamic systems and in handling over-parametrization and collinearity have made them a popular choice in the field of nowcasting, see for instance Bańbura et al. (2010), Cimadomo et al. (2020), or Schorfheide and Song (2015).

For this analysis, the *estimate_mfbvar* function of the *mfbvar* (Ankargren et al., 2021) R library was used to estimate and predict on a Bayesian VAR model. A Minnesota prior coupled with the inverse Wishart prior for the form of the error variance-covariance matrix was used, as performed in Cimadomo et al. (2020).

### 2.2c Decision tree

The decision tree is a commonly used, non-parametric algorithm in machine learning, due in part to its simplicity and interpretability. Decision trees are often employed as part of an ensemble approach, combining many decision trees as weak learners to produce a strong learner. Two of those tree-based ensemble approaches, gradient boosted trees and random forest, will be examined in the coming sections. The basic premise of a decision tree does not differ from the standard semantic interpretation of the term; all data begin as one group at the "root" of the tree and are then split into "branches" at different nodes depending on their characteristics and the information gain from that split. This splitting can be very general, with for instance only a single split, or, at its most extreme, continuing until every observation sits alone on its own "leaf". Decision trees are normally not equipped to handle time series data, see section four for more information on how this was addressed for the decision tree and other methodologies which do not natively handle time series. Simple decision trees have not been used for nowcasting, though tree-based ensembles in nowcasting were examined in Soybilgen and Yazgan (2021) or Tiffin (2016). For more information on decision trees, see Patel and Prajapati (2018) or scikit-learn (2021a).

Decision trees have hyperparameters which usually need to be tuned depending on the application. In machine learning, hyperparameter refers to parameters which determine the macrostructure of a model and not the model's coefficients and parameters themselves. An example with decision trees is the max depth of the tree, or the number of splits the tree can have, which is then a given condition of the structure of the model independent of the data used to train it. Coefficients within the model, i.e., how to split the data, are then determined from the training data the model is fit with. Hyperparameter tuning refers to the process of establishing a performance metric, e.g., mean absolute error (MAE) or root mean square error (RMSE) in a regression application, testing out different hyperparameter combinations, recording their performance according to the performance metric, and selecting a final value for the algorithm's hyperparameters. For more information on hyperparameter tuning, see Probst et al. (2018).

For this analysis, the *DecisionTreeRegressor* function of the *sklearn* (scikit-learn, 2021b) Python library was used. See section four for more information on how hyperparameters were selected.

## 2.2d Dynamic factor model

Dynamic factor models (DFM) are commonly used in time series forecasting and nowcasting. They operate under the assumption that one or several latent underlying factors explain the development of multiple time series, which are a product of these latent factors plus an idiosyncratic error term. The latent factor could represent, for instance, the business cycle. DFMs are often estimated by assigning variables to various "blocks", which can represent different aspects of the economy. For example, when nowcasting GDP, all variables can be assigned to one global block, while a subset of variables can be assigned to another block, representing, e.g., a geography or economic sector. For more information on the use of blocks in estimating DFMs, see Hallin and Liška (2011).

DFMs are one of the most commonly applied methodologies in nowcasting. Some examples include nowcasting Canadian GDP growth (Chernis and Sekkel, 2017), global trade growth (Cantú, 2018; Guichard and Rusticelli, 2011), Russian GDP growth (Porshakov et al., 2016), and German GDP growth (Marcellino and Schumacher, 2010), among many others. For more information on DFMs, see any of Bok et al. (2018), Cantú (2018), Giannone et al. (2005), or Stock and Watson (2002).

For this analysis, the *dfm* function of the *nowcastDFM* (Hopp and Cantú, 2020) R library was used. This library was developed as an R implementation of the original MATLAB code published alongside Bok et al. (2018). This implementation of the DFM is modelled in state-space form under the assumption that all variables share common latent factors in addition to their own idiosyncratic components. Subsequently, the Kalman filter is applied and parameter estimates are obtained via maximum likelihood estimation. For more information on this particular modelling approach, see Bańbura and Rünstler (2011), Bok et al. (2018), and Cantú (2018). Blocks used were the same specified in Bok et al. (2018). Using a single, global block was also assessed, but was found to achieve significantly worse results than Bok et al. (2018)'s block specification.

## 2.2e Gradient boosted trees

Gradient boosted trees is an ensemble machine learning algorithm combining the result of several simpler decision tree models, similar to the random forest method discussed in section 2.2k. Gradient boosted trees combines individual decision trees sequentially, with each addition trained to reduce the errors of the previous iteration. The modelling approach has been shown to be powerful and high-performing in a variety of applications, counting as one of the most commonly winning algorithms in the Kaggle data science competition (Boehmke, 2018). For more information on the gradient boosting approach, see Natekin and Knoll (2013).

For this analysis, the *GradientBoostingRegressor* function of the *sklearn* (scikit-learn, 2021b) Python library was used. See section four for more information on how hyperparameters were selected.

## 2.2f Long short-term memory artificial neural network

Long short-term memory artificial neural networks (LSTM) are a sub-class of artificial neural network (ANN). For more information on ANNs, see section 2.2i. The traditional feedforward network discussed in that section has information flowing through it unidirectionally, from the beginning of the network to the end. Recurrent neural networks (RNN) were introduced to make ANNs more suitable for time series and situations where there is a temporal dependency in the data, for instance in applications like speech processing. RNNs introduce a feedback loop to the traditional ANN architecture, where outputs of layers can be fed back into the network before a final output is obtained. For more information on RNNs, see Dematos et al. (1996) or Stratos (2020). Due to vanishing and exploding gradients, RNNs tend to have a short memory and thus are of limited use in the nowcasting context. For more information on vanishing and exploding gradients in RNNs, see Grosse (2017). LSTMs address this deficiency by introducing a memory cell and an input, output, and forget gate. Gradients are then able to flow through the network unchanged, allowing LSTMs to establish longer time dependencies than RNNs. For more information on LSTM architecture, see Brownlee (2018) or Chung et al. (2014).

Though LSTMs have been used before for nowcasting meteorological events (Shi et al., 2015), their fitness for economic nowcasting has only recently begun to be explored. One recent United Nations Conference on Trade and Development (UNCTAD) research paper examined their suitability in nowcasting global trade, finding them to produce superior results to DFMs (Hopp, 2021a). An open-source library for nowcasting economic data using LSTMs in multiple programming languages was published alongside this paper. A second UNCTAD research paper compared their performance with the DFM in nowcasting global trade during the COVID crisis, again with the LSTM more often producing superior performance (Hopp, 2021b).

For this analysis, the *LSTM* function of the *nowcast_lstm* (Hopp, 2021c) Python library was used. See section four for more information on how hyperparameters were selected.

## 2.2g Mixed-data sampling regression

Introduced in 2004, the MIDAS framework was developed to address the issue of estimating time series regression models where the dependent variable is sampled at a lower frequency than the independent variables (Ghysels et al., 2004). Similar to mixed-frequency VAR, MIDAS can suffer from over-parametrization due to the necessity of including the lags of higher frequency variables in the model. To avoid this issue and obtain parsimony, MIDAS regression frequently employs a non-linear weighting scheme for the lag coefficients (Kuzin et al., 2011). MIDAS' ability to handle mixed frequency data as well as mitigate over-parametrization make them well-suited for nowcasting, where the methodology has been employed for instance in nowcasting Euro area GDP (Kuzin et al., 2009) or German GDP (Marcellino and Schumacher, 2010).

For this analysis, the *midas_r* function of the *midasr* (Kvedaras, 2021) R library was used employing an exponential Almon weighting function for lag coefficients. As specified in Kuzin et al. (2009), separate univariate models were estimated for each independent variable whose forecasts were then combined via a weighted mean. Weights were found using RMSE of the model on the training set, adjusted to discount the worst performing univariate model, so that this model had no contribution to the combined forecast. See Clements and Galvão (2008) for more information on forecast combination in MIDAS models.

## 2.2h Mixed-frequency vector autoregression

As stated in section 2.2b, vector autoregression estimates a set or vector of variables as a function of the vector's own prior $p$ values. Originally introduced in 1980 (Sims, 1980), VAR models have become popular in the domain of macroeconomic forecasting due to their ability to capture the dynamics of complex systems using multiple time series (Stock and Watson, 2001). They do, however, come with several limitations which limit their usefulness in the nowcasting context. Firstly, they do not natively handle mixed-frequency data. Secondly, they often suffer from over-parametrization, where the inclusion of additional variables leads to exponentially more parameters (Kuzin et al., 2009). The first issue can be addressed by three different approaches. One common approach is aggregating higher frequency indicators to the frequency of the lowest frequency indicator. Taking the case of estimating a VAR for a quarterly growth rate using monthly variables, the monthly variables can be transformed to a quarterly frequency by for instance averaging the growth rates for the three constituent months, or by calculating the full quarterly growth rate from the monthly values. A second approach is stacking the constituent months of a quarter into three separate series, so that one monthly series becomes three quarterly series composed of the time series for months one, two, and three of each quarter. A final approach is estimating the model in the highest frequency of the dataset and interpolating lower-frequency variables into this higher frequency (Ghysels, 2016). Any of these approaches yields a mixed-frequency VAR model (MF-VAR).

MF-VAR models have been used in nowcasting applications before, for instance for nowcasting Euro area GDP (Kuzin et al., 2009). However, due to the aforementioned issues with over-parametrization, Bayesian VARs are more often applied. See section 2.2b for more information on how Bayesian VARs help address this issue.

For this analysis, the *VAR* function of the *PyFlux* (Taylor, 2016) Python library was used to estimate and predict the MF-VAR model.

## 2.2i Multilayer perceptron feedforward artificial neural network

Artificial neural networks (ANN) have become extremely popular in recent years due to their exceptional performance in a variety of applications, ranging from image recognition to speech processing to self-driving cars. ANNs are made up of inter-connected layers of nodes which receive data inputs, either from an external data source or from a previous layer, run the data through a series of weights or coefficients and a non-linear activation layer, then generate an output. This output can either be the final prediction of the model or serve as an input to a future layer. The model is trained by defining a cost function, calculating gradients with respect to this cost function, then adjusting weights in the direction of minimizing error according to the cost function. This process is called back propagation. The most common type of ANN is the feedforward multilayer perceptron (MLP), where information flows unidirectionally through the network. For more information on ANNs, see Sazli (2006) or Singh and Prajneshu (2008).

ANNs have been used for economic and forecasting applications, with good results. Examples include nowcasting US GDP (Loermann and Maas, 2019), forecasting commodity prices (Kohzadi et al., 1996), and forecasting exchange rates (Falat and Pancikova, 2015).

For this analysis, the *MLPRegressor* function of the *sklearn* (scikit-learn, 2021b) Python library was used to estimate and predict with an MLP network.

## 2.2j Ordinary least squares regression

Ordinary least squares is one of the most commonly employed approaches in regression applications thanks to its simplicity, power, and interpretability. OLS estimates the parameters of a linear function mapping a set of input variables to an output variable. These values are determined by minimizing the sum of squared differences between the actual output variable and the estimate of the linear function. See B. et al. (2018) for more information on OLS.

Generally, OLS is not well-suited for use in the nowcasting context as it contains no explicit provision for time series and often suffers from multicollinearity when higher numbers of input variables are used, rendering its coefficients unstable and violating a base assumption of the model. See section four for more information on the adjustments to the data required to use OLS in this analysis.

For this analysis, the *LinearRegression* function of the *sklearn* (scikit-learn, 2021b) Python library was used to estimate and predict with the OLS model.

## 2.2k Random forest

Random forest is an ensemble machine learning algorithm that combines the results of multiple decision tree models. See section 2.2d for more information on decision trees. Individual decision trees often have poor performance on their own, tending to overfit their training data if allowed too many layers. Random forest trains many decision trees on different subsets of the training data and averages their predictions to obtain more generalizable estimates with lower variance. Random forest models are not generally used in nowcasting, but were examined in Soybilgen and Yazgan (2021) and Tiffin (2016).

For this analysis, the *RandomForestRegressor* function of the *sklearn* (scikit-learn, 2021b) Python library was used to estimate and predict with a random forest model.

### 2.2l Ridge regression

Ridge regression is very similar to the OLS approach examined in section 2.2j. However, it seeks to address one of the principal issues with OLS: multicollinearity. Ridge regression adds regularization to OLS through the introduction of a penalty term. The model then seeks to minimize not only the residual sum of squares, but also this penalty term. In this way, coefficients more often tend towards zero. Ridge regression allows the inclusion of more input variables than standard OLS, an important characteristic for nowcasting. This penalty term necessitates the selection of one hyperparameter at estimation, an alpha term denoting the strength or degree of regularization. For more information on ridge regression in the context of nowcasting, see Tiffin (2016).

For this analysis, the *RidgeRegression* function of the *sklearn* (scikit-learn, 2021b) Python library was used to estimate and predict with a ridge regression model.

## 3. Data

Data for US GDP, the target variable, and all explanatory variables were gotten from the Federal Reserve Economic Data (FRED) API (Boysel and Vaughan, 2021). Explanatory variables were those specified in Bok et al. (2018), encompassing a variety of monthly and quarterly economic indicators such as goods and services exports, building permits, various price indices, and retail sales, among others. All series were obtained from FRED already seasonally adjusted, where applicable. All series were then transformed to period over period growth rates. Data for GDP ranges from the first quarter of 1947 to the third quarter of 2021. Availability start dates for explanatory variables range from January 1947 to November 2009.

Before proceeding, it is necessary to define the terms training, validation, and test periods or sets. These terms refer to an important concept in model evaluation, and machine learning in general, of assessing model performance on data the model was not trained on. The training set refers to the data used to train or estimate the model, while the test set refers to the data the trained model's performance is ultimately evaluated on. The purpose of splitting the data in these sets is to ensure the generalizability of the model and to avoid overfitting. A model may perform very well on predicting values from the data it was trained on, but may generalize very poorly to new data it has not seen before. The validation set is similar to the test set, but is used exclusively in selecting hyperparameters for algorithms that require them. The idea is to further divide the training set into a validation set so that different hyperparameters can be tested and selected based on their performance in predicting this validation set. Once hyperparameters are selected, the model can be retrained with the full training set and finally assessed on the test set. The logic of utilizing a separate validation and test set is to avoid information leakage and thus overfitting to the test set. A model with hyperparameters selected based on performance on the test set may be overfit to that set and not generalize well. For more information on these concepts, see scikit-learn (2021a).

Three separate test periods were used for assessing the performance of the models, each representing a volatile period in US economic history. This helped to ensure the robustness of results, testing models across very disparate eras with large economic shocks. The first test period dated from the first quarter of 1972 to the fourth quarter of 1983. This period spanned three different US recessions, one in 1974, one in 1980, and one in 1982 (Sablik, 2013), making it ideal to test the robustness of model results. This test period will hereafter be referred to as either "period 1", or "early 1980s recession". The second test period dated from the first quarter of 2005 to the fourth quarter of 2010. This period encompassed the 2008 financial crisis. This period will hereafter be referred to as "period 2", or "financial crisis". The final test period dated from the first quarter of 2016 to the third quarter of 2021, encompassing four years before the COVID crisis and everything up to the present day. This period will hereafter be referred to as "period 3", or "COVID crisis". The training periods for each test period corresponded to the first quarter of 1947 to the fourth quarter of 1971 for period 1, the first quarter of 1947 to the fourth quarter of 2004 for period 2, and the first quarter of 1947 to the fourth quarter of 2015 for period 3. For methodologies which required hyperparameter tuning, the validation set corresponded to the first quarter of 1966 to the fourth quarter of 1971 for period 1, the first quarter of 1992 to the fourth quarter of 2004 for period 2, and the first quarter of 2006 to the fourth quarter of 2015 for period 3.

Of the variables specified in Bok et al. (2018), not all had a long enough time series to be included in the earlier test and validation periods. As a result, only variables with a time series dating back to the fourth quarter of 1960, to the fourth quarter of 1991, and to the fourth quarter of 2005 were included in the period 1, 2, and 3 models, respectively.

# 4. Modelling approach

To obtain the results in section five, a model for each methodology outlined in section two was fit using the training set, consisting of data up to the train end dates specified in the previous section. Model predictions were then obtained on the test set for each test period at five different simulated data vintages. A simulated data vintage refers to the artificial introduction of missing values into historical data to simulate the data as it would have appeared at different points in the past. Testing model performance at different time points is important, as series in a live nowcasting model will rarely have complete data for the quarter at the time of prediction. A nowcast for a given quarter will most likely begin in media res, or even before the quarter has begun. A nowcasting model must be robust to this lack of information earlier in the prediction period. The five data vintages examined were two months before the prediction period, one month before the prediction period, the month of the prediction period, the month after the prediction period, and two months after the prediction period. If the nowcast is for the second quarter, two months before the prediction period refers to the data as it would have appeared in April, and so on. GDP actuals are ostensibly already available two months after the quarter, but this value is an advanced estimate liable to revision, so it is worth evaluating the models even at this advanced time point. Information for publication lags of each of the variables was obtained from Bok et al. (2018).

For methodologies with hyperparameters, the validation set for each test period outlined in the previous section was used to determine those hyperparameters. For the final hyperparameters and full code used to estimate each model, see Hopp (2022).

It is worth noting that the three autoregressive methodologies, namely ARMA, Bayesian VAR, and MF-VAR, do not strictly adhere to the training set-test set paradigm. In all other methodologies, only information from the training set is used to fit the models and estimate parameters. The models are then fixed and new data from the test set is only used for inference. In the implementations for the autoregressive methodologies, this is not possible. Rather, they are estimated on a dataset and then generate one or two-step ahead predictions. This means that newer information dating from beyond the training end date is used in estimating their coefficients. For the other methodologies, this would be the equivalent of retraining the models with all the data available up until the prediction period rather than only until the training end date. The autoregressive models then have access to slightly more information in the results presented in section five, but due to the decades-long span of the training sets, this advantage is likely to be minimal. Even still, it is worth mentioning.

Another characteristic of this dataset is that not all series have a publication history as long as GDP's. This raises the question of how to handle these series in training the model. There are three approaches that can be taken. The first approach is to simply drop any series with a start date later than the first quarter of 1947, the start date of our target variable. This is an approach that drops columns or series from our data. The second is to drop any observations earlier than the latest start date of all the series. This is an approach that drops rows from our data. The final approach is imputing the missing values using various techniques, such as mean imputation or expectation maximization (EM) (Soley-Bori, 2013). For this analysis, the third approach, filling missing values with the mean, was employed, though all three approaches were tested empirically.

The first, dropping variables with a start date later than the first quarter of 1947, left only five explanatory variables out of the 28 specified in Bok et al. (2018). Performance was consequently worse across all methodologies. The second, dropping observations before the latest variable start date in the dataset, also obtained worse performance across all methodologies. That left the third option, along with the decision of how to fill missing values. Two of the approaches outlined in Soley-Bori (2013) were tested: filling values with the mean, and filling them using EM imputation. The two approaches yielded comparable results, with some methodologies performing slightly better with one approach and others with the other. The decision to use mean-filling was then taken based on two factors: the aforementioned comparable empirical performance, and for consistency with the filling ragged-edges approach outlined below.

Ragged-edges refer to missing values at the ends of time series in the data due to differing publication schedules and nowcasting at different times throughout the prediction period. For instance, a nowcast for the second quarter performed in May must have all values for June missing, unless the variable is some kind of forecast or plan. There are many different approaches to dealing with ragged-edges, and three were tested for this analysis. The first, ultimately chosen, was filling missing values with the series mean. The second was filling missing values with n-step ahead forecasts of univariate ARMA models for each series. The third was using EM imputation. All three approaches had similar empirical performance, again with certain methodologies performing better with one approach and others with a different one, but differences were marginal. The decision to use mean-filling was then taken based on two considerations: first, the aforementioned parity in performance, and a second, practical consideration. Filling missing values with the mean is a very light computational task, occurring nearly instantaneously. EM imputation is a complex quantitative approach whose calculation time scales with both the number of observations and the number of variables. Because each data vintage, that is, combination of target period and simulated lag, has a unique pattern of missing values, this requires a run of the algorithm for each one. This quickly becomes computationally expensive and impractical for assessing model performance. Imputing ragged-edges on one data vintage for the third test period via the EM method takes roughly 3.5 minutes, for instance. This needs to be performed for 23 quarters and five lags, resulting in 115 different data vintages. This translates to a run time of more than 8 hours. This process needs to be rerun each time a variable is added or removed from the model, a frequent occurrence in the model selection process. The situation is similar, though not as severe, for ARMA ragged-edge filling. An ARMA model needs to be fit for each data vintage, though need not be rerun if variables are added or removed from the model, as the series models are univariate. Consequently, mean-filling was selected for dealing with ragged-eges for all methodologies, improving comparability. The sole exception was the DFM, where use of EM imputation in combination with the Kalman filter is an essential component of the implementations in Bok et al. (2018) and Hopp and Cantú (2020)

The final data issue to address in the analysis was how to handle mixed frequency time series for the methodologies which do not natively handle them. As described in section 2.2h, there are three approaches that can be taken. Higher frequency indicators can be aggregated to a lower frequency, they can be stacked into separate series, or lower frequency indicators can be converted to a higher frequency, with missing values interpolated. For this analysis, the first two approaches were tested, as a good implementation of the interpolation method detailed in Kuzin et al. (2011) could not be found. For the first approach, two methods of aggregation were tested: taking the average growth rate of the three months constituting a quarter, and calculating the full quarterly growth rate via each individual month's growth rate. Stacking monthly series was found to perform better than aggregating across methodologies, so was the approach utilized. Methodologies where stacking the time series was relevant were the decision tree, gradient boosted trees, MF-VAR, MLP, OLS, ridge regression, and random forest.

# 5. Results

Figures 1, 2, and 3 show model predictions for each of the three test periods at different data vintages. The y-axis displays quarterly growth rate, the black line displays the actual observed quarterly growth rate, while the various colored dotted lines represent nowcasts at different data vintages. Methodologies are ordered by best performance according to RMSE, left to right, top to bottom.

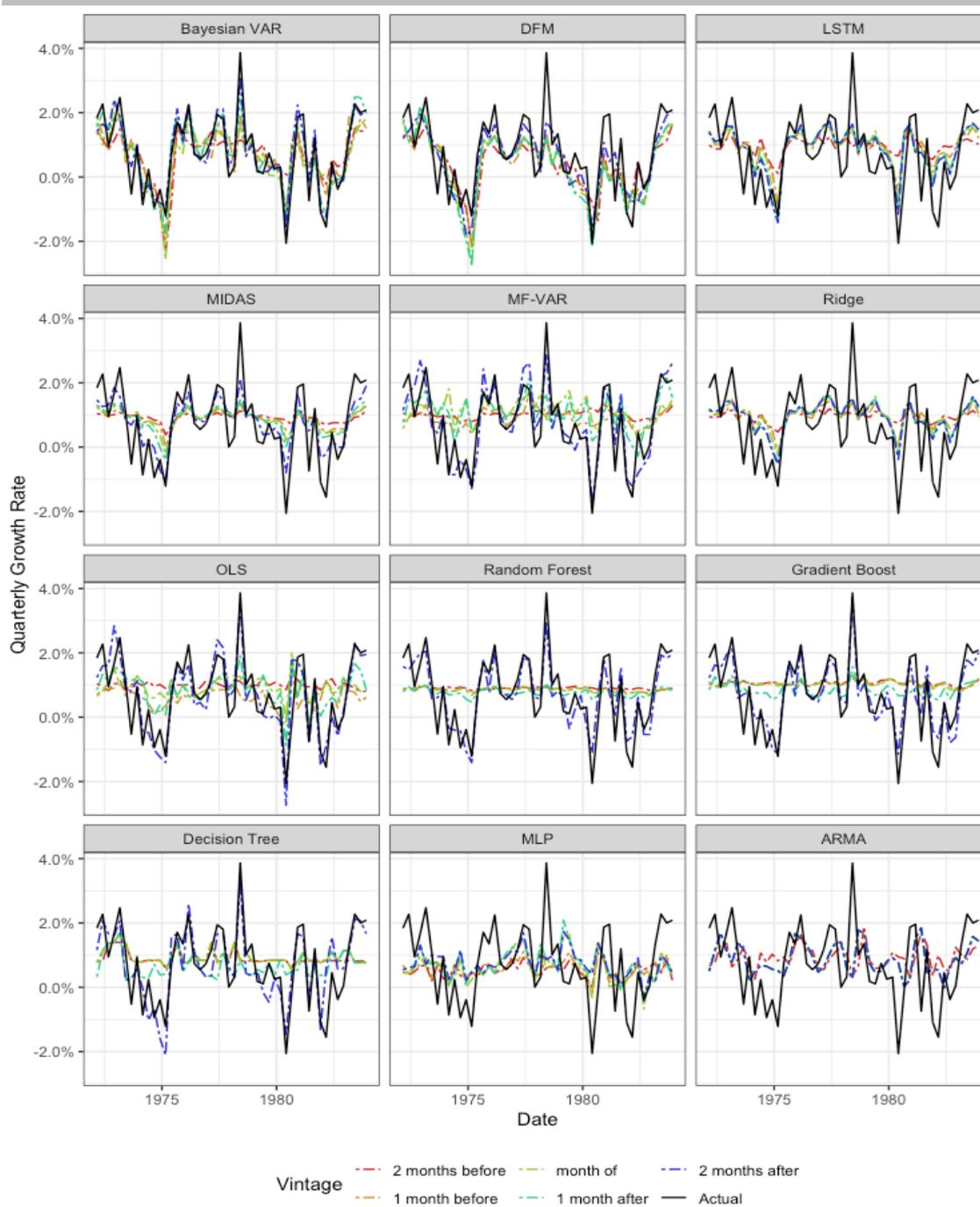

Figure 1. Nowcasts for test period 1: early 80s recession

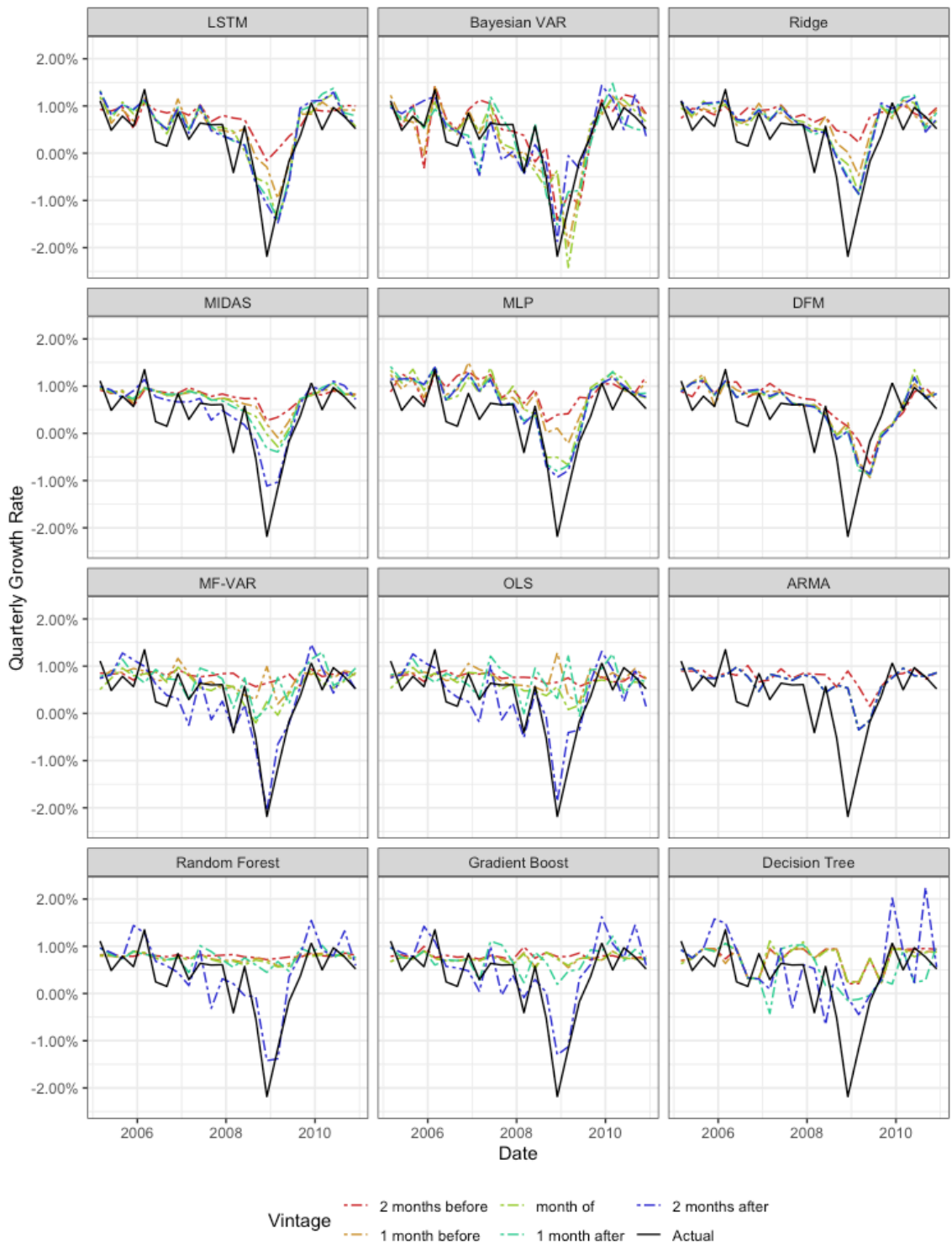

**Figure 2.** Nowcasts for test period 2: financial crisis

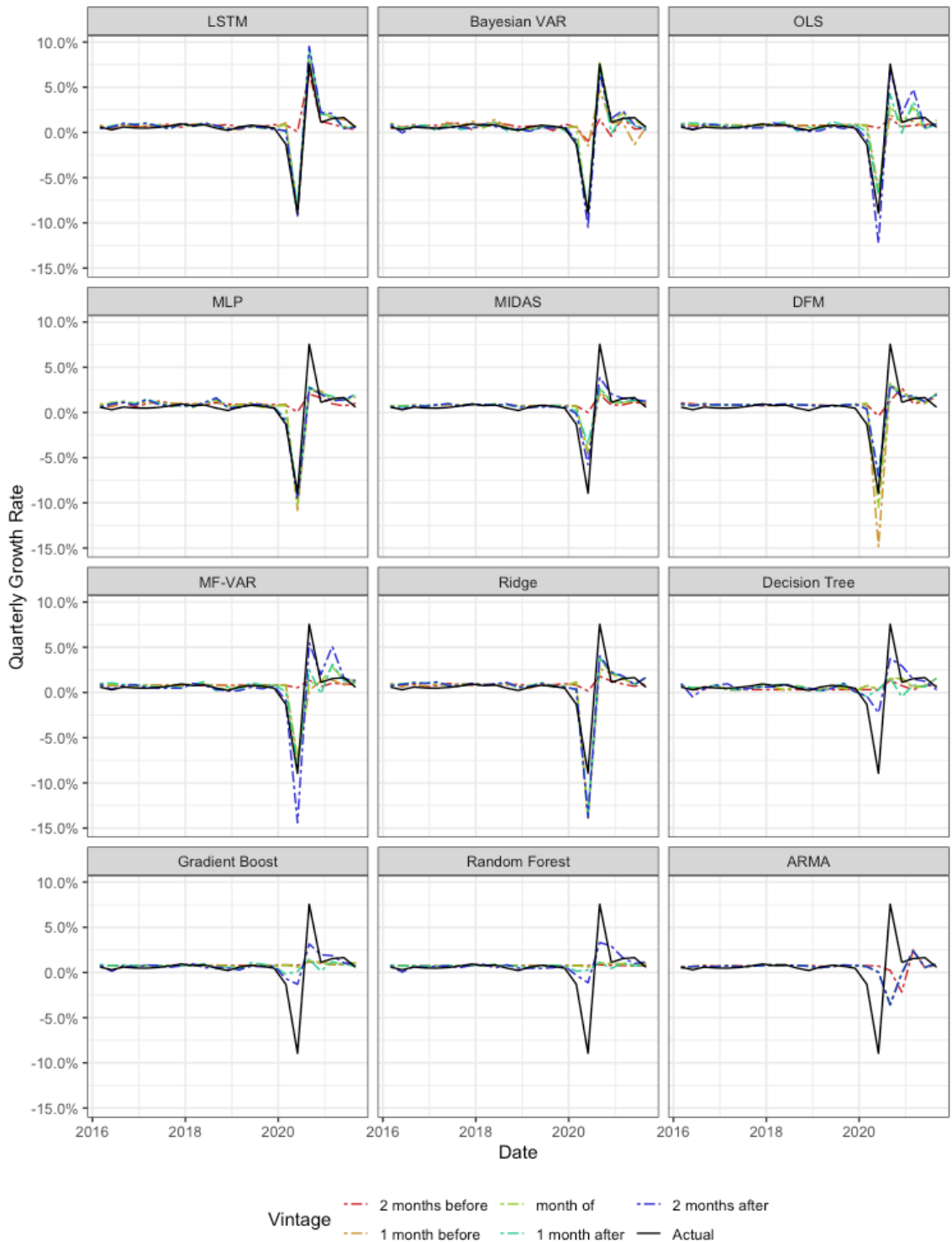

**Figure 3.** Nowcasts for test period 3: COVID crisis

Tables 1-6 display MAE and RMSE of each methodology as a proportion of the benchmark ARMA model's MAE and RMSE.

### Table 1. Test period 1: early 80s financial crisis, MAE as a proportion of ARMA model

| Vintage | ARMA | Bayesian VAR | Decision Tree | DFM | Gradient Boost | LSTM | MF-VAR | MIDAS | MLP | OLS | Random Forest | Ridge |
|---|---|---|---|---|---|---|---|---|---|---|---|---|
| 2 months before | 1 | 0.723 | 0.954 | 0.736 | 0.987 | 0.896 | 0.958 | 0.913 | 0.978 | 1.011 | 0.982 | 0.926 |
| 1 month before | 1 | 0.623 | 0.979 | 0.662 | 1.010 | 0.772 | 0.913 | 0.851 | 0.976 | 1.007 | 0.987 | 0.873 |
| month of | 1 | 0.699 | 0.981 | 0.714 | 0.997 | 0.730 | 1.002 | 0.810 | 1.014 | 1.010 | 0.977 | 0.862 |
| 1 month after | 1 | 0.374 | 0.925 | 0.737 | 0.895 | 0.696 | 0.809 | 0.750 | 1.041 | 0.784 | 0.934 | 0.820 |
| 2 months after | 1 | 0.335 | 0.563 | 0.634 | 0.442 | 0.669 | 0.456 | 0.518 | 0.999 | 0.452 | 0.457 | 0.790 |
| Average | 1 | 0.552 | 0.881 | 0.697 | 0.867 | 0.754 | 0.829 | 0.770 | 1.001 | 0.854 | 0.868 | 0.855 |

### Table 2. Test period 2: financial crisis, MAE as a proportion of ARMA model

| Vintage | ARMA | Bayesian VAR | Decision Tree | DFM | Gradient Boost | LSTM | MF-VAR | MIDAS | MLP | OLS | Random Forest | Ridge |
|---|---|---|---|---|---|---|---|---|---|---|---|---|
| 2 months before | 1 | 0.732 | 1.049 | 0.910 | 1.090 | 0.857 | 1.004 | 0.950 | 1.136 | 1.002 | 1.043 | 1.032 |
| 1 month before | 1 | 0.830 | 1.189 | 0.961 | 1.153 | 0.692 | 1.104 | 0.950 | 1.050 | 1.133 | 1.116 | 0.897 |
| month of | 1 | 0.828 | 1.166 | 0.933 | 1.153 | 0.686 | 0.952 | 0.902 | 0.903 | 0.983 | 1.071 | 0.822 |
| 1 month after | 1 | 0.660 | 1.119 | 0.911 | 1.086 | 0.703 | 1.040 | 0.820 | 0.817 | 1.169 | 1.106 | 0.814 |
| 2 months after | 1 | 0.800 | 1.255 | 0.924 | 0.781 | 0.683 | 0.699 | 0.638 | 0.752 | 0.802 | 0.837 | 0.780 |
| Average | 1 | 0.769 | 1.153 | 0.928 | 1.054 | 0.727 | 0.961 | 0.854 | 0.936 | 1.018 | 1.035 | 0.873 |

### Table 3. Test period 3: COVID crisis, MAE as a proportion of ARMA model

| Vintage | ARMA | Bayesian VAR | Decision Tree | DFM | Gradient Boost | LSTM | MF-VAR | MIDAS | MLP | OLS | Random Forest | Ridge |
|---|---|---|---|---|---|---|---|---|---|---|---|---|
| 2 months before | 1 | 0.858 | 0.882 | 0.920 | 0.854 | 0.631 | 0.843 | 0.782 | 0.876 | 0.841 | 0.862 | 0.840 |
| 1 month before | 1 | 0.690 | 0.862 | 0.730 | 0.803 | 0.251 | 0.579 | 0.564 | 0.631 | 0.573 | 0.811 | 0.678 |
| month of | 1 | 0.215 | 0.878 | 0.550 | 0.802 | 0.346 | 0.522 | 0.567 | 0.612 | 0.524 | 0.810 | 0.656 |
| 1 month after | 1 | 0.273 | 0.856 | 0.537 | 0.783 | 0.331 | 0.550 | 0.564 | 0.489 | 0.582 | 0.779 | 0.613 |
| 2 months after | 1 | 0.306 | 0.665 | 0.539 | 0.633 | 0.359 | 0.587 | 0.444 | 0.487 | 0.475 | 0.686 | 0.623 |
| Average | 1 | 0.467 | 0.828 | 0.654 | 0.775 | 0.383 | 0.615 | 0.583 | 0.618 | 0.598 | 0.790 | 0.681 |

### Table 4. Test period 1: early 80s financial crisis, RMSE as a proportion of ARMA model

| Vintage | ARMA | Bayesian VAR | Decision Tree | DFM | Gradient Boost | LSTM | MF-VAR | MIDAS | MLP | OLS | Random Forest | Ridge |
|---|---|---|---|---|---|---|---|---|---|---|---|---|
| 2 months before | 1 | 0.718 | 0.936 | 0.731 | 0.973 | 0.878 | 0.948 | 0.897 | 0.937 | 0.974 | 0.957 | 0.900 |
| 1 month before | 1 | 0.622 | 0.972 | 0.675 | 1.008 | 0.775 | 0.918 | 0.850 | 0.942 | 0.981 | 0.969 | 0.855 |
| month of | 1 | 0.681 | 0.974 | 0.736 | 0.995 | 0.750 | 1.028 | 0.807 | 0.990 | 0.982 | 0.961 | 0.840 |
| 1 month after | 1 | 0.398 | 0.909 | 0.757 | 0.876 | 0.708 | 0.825 | 0.754 | 1.031 | 0.779 | 0.913 | 0.804 |
| 2 months after | 1 | 0.331 | 0.566 | 0.641 | 0.437 | 0.686 | 0.457 | 0.540 | 0.997 | 0.468 | 0.442 | 0.778 |
| Average | 1 | 0.552 | 0.872 | 0.708 | 0.859 | 0.760 | 0.836 | 0.771 | 0.979 | 0.838 | 0.849 | 0.836 |

### Table 5. Test period 2: financial crisis, RMSE as a proportion of ARMA model

| Vintage | ARMA | Bayesian VAR | Decision Tree | DFM | Gradient Boost | LSTM | MF-VAR | MIDAS | MLP | OLS | Random Forest | Ridge |
|---|---|---|---|---|---|---|---|---|---|---|---|---|
| 2 months before | 1 | 0.573 | 0.915 | 0.817 | 1.056 | 0.760 | 0.987 | 0.892 | 0.971 | 0.985 | 1.014 | 0.916 |
| 1 month before | 1 | 0.734 | 1.069 | 0.918 | 1.167 | 0.675 | 1.110 | 0.931 | 0.927 | 1.196 | 1.128 | 0.824 |
| month of | 1 | 0.778 | 1.056 | 0.877 | 1.163 | 0.626 | 0.908 | 0.851 | 0.762 | 0.990 | 1.103 | 0.741 |
| 1 month after | 1 | 0.538 | 0.917 | 0.849 | 1.007 | 0.577 | 0.982 | 0.758 | 0.690 | 1.127 | 1.070 | 0.693 |
| 2 months after | 1 | 0.627 | 1.078 | 0.858 | 0.594 | 0.550 | 0.521 | 0.521 | 0.641 | 0.574 | 0.632 | 0.678 |
| Average | 1 | 0.648 | 1.004 | 0.862 | 0.999 | 0.641 | 0.904 | 0.794 | 0.803 | 0.975 | 0.990 | 0.775 |

**Table 6. Test period 3: COVID crisis,** RMSE as a proportion of ARMA model

| Vintage | ARMA | Bayesian VAR | Decision Tree | DFM | Gradient Boost | LSTM | MF-VAR | MIDAS | MLP | OLS | Random Forest | Ridge |
|---|---|---|---|---|---|---|---|---|---|---|---|---|
| 2 months before | 1 | 0.816 | 0.892 | 0.865 | 0.929 | 0.737 | 0.904 | 0.842 | 0.854 | 0.896 | 0.940 | 0.866 |
| 1 month before | 1 | 0.608 | 0.788 | 0.561 | 0.802 | 0.191 | 0.543 | 0.490 | 0.446 | 0.467 | 0.813 | 0.526 |
| month of | 1 | 0.153 | 0.785 | 0.384 | 0.804 | 0.219 | 0.492 | 0.493 | 0.416 | 0.406 | 0.813 | 0.482 |
| 1 month after | 1 | 0.140 | 0.777 | 0.382 | 0.762 | 0.202 | 0.401 | 0.517 | 0.372 | 0.348 | 0.783 | 0.433 |
| 2 months after | 1 | 0.175 | 0.555 | 0.383 | 0.614 | 0.220 | 0.482 | 0.358 | 0.377 | 0.345 | 0.632 | 0.457 |
| Average | 1 | 0.368 | 0.756 | 0.507 | 0.779 | 0.303 | 0.556 | 0.533 | 0.484 | 0.483 | 0.793 | 0.545 |

Tables 7-9 display the average revision between two data vintages of each methodology, ordered from least revision to most. In table 7, for instance, we can see that when moving from one data vintage to another, i.e., more data is available to the model, ridge regression revised its nowcast by 0.09 percentage points, while OLS' revisions were on average almost five times larger, at 0.42 percentage points. An ideal model would have low values in tables 1-6, indicating good predictive performance, and low values in tables 7-9, indicating lower month-to-month revisions and more consistent predictions.

**Table 7. Test period 1: early 80s recession,** average month-to-month revision

| Method | Average revision |
|---|---|
| ARMA | 0.07% |
| Ridge | 0.09% |
| LSTM | 0.12% |
| MIDAS | 0.14% |
| MLP | 0.16% |
| Random Forest | 0.22% |
| DFM | 0.23% |
| Gradient Boost | 0.23% |
| Decision Tree | 0.25% |
| Bayesian VAR | 0.34% |
| MF-VAR | 0.38% |
| OLS | 0.42% |

**Table 8. Test period 2: financial crisis,** average month-to-month revision

| Method | Average revision |
|---|---|
| ARMA | 0.04% |
| DFM | 0.07% |
| MIDAS | 0.1% |
| Ridge | 0.11% |
| MLP | 0.14% |
| LSTM | 0.15% |
| Random Forest | 0.16% |
| Gradient Boost | 0.17% |
| Decision Tree | 0.19% |
| OLS | 0.27% |
| MF-VAR | 0.28% |
| Bayesian VAR | 0.3% |

**Table 9. Test period 3: COVID crisis,** average month-to-month revision

| Method | Average revision |
|---|---|
| ARMA | 0.09% |
| Random Forest | 0.17% |
| MIDAS | 0.18% |
| Gradient Boost | 0.18% |
| Decision Tree | 0.27% |
| LSTM | 0.3% |
| Ridge | 0.32% |
| MLP | 0.32% |
| DFM | 0.33% |
| OLS | 0.43% |
| MF-VAR | 0.47% |
| Bayesian VAR | 0.52% |

With three different test periods, five different data vintages, two different performance metrics, and secondary performance metrics like average revision, it can be difficult to make sense of the results presented above. Figure 4 attempts to summarize all of these results in a more digestible manner by producing an overall aggregate score for each methodology, scaled to between zero and one. It was calculated in the following manner: for each test period, MAE and RMSE were individually averaged across all vintages. The two resulting figures were then in turn averaged, leaving one number for each methodology in each time period. The figures for each time period were then scaled to be between zero and one, with zero being the best performing methodology in the time period, and one being the worst performing. To account for the degree of volatility in predictions, the values from tables 7-9 were also scaled to be between zero and one, with the least-revised model in a time period scaled to zero and the most-revised model scaled to one. To account for the fact that revision volatility was only a secondary performance metric in the model, the scaled score for the three test periods was averaged to obtain a single figure for each methodology. For each methodology, this figure was then added to the three figures obtained from the MSE and RMSE transformation outlined above. This left four scaled performance metrics per methodology: one for test period 1, one for test period 2, one for test period 3, and one for revision volatility. These four figures were then averaged to obtain the values in figure 4. A lower value on the chart indicates a better-performing model across all test periods and vintages in terms of MAE, RMSE, and revision volatility. A model with a value of zero (one) would indicate a model that performed the best (worst) in all five data vintages for all three test periods while also revising its predictions the least (most) across all those same vintages and test periods.

**Figure 4.** Overall aggregated and scaled methodology performance (lower is better)

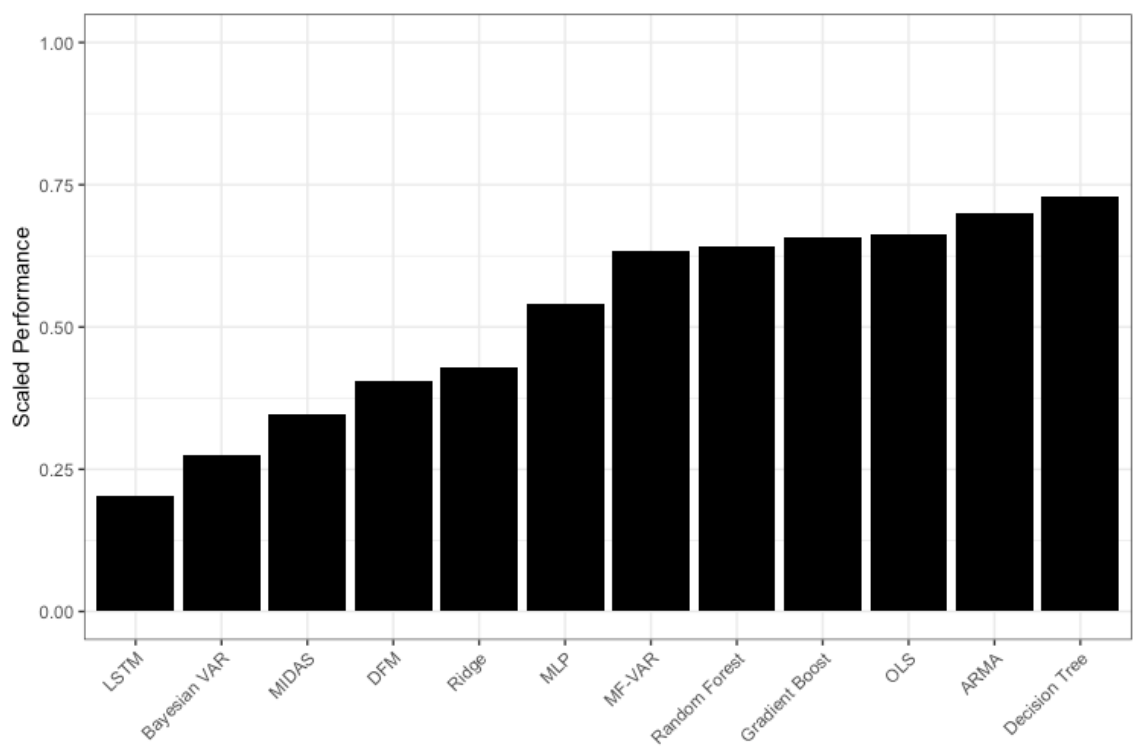

In terms of this aggregated performance metric, the LSTM was the best-performing methodology in this analysis. It was then followed by Bayesian VAR, MIDAS, and the DFM, with performance decreasing in roughly equal increments for each, while ridge regression performed closely to the DFM. MLP followed with a score close to equidistant between ridge regression and MF-VAR. MF-VAR, random forest, gradient boost, and OLS performed similarly poorly, while ARMA and the decision tree rounded out the rear. Detailed observations for each methodology follow below.

### ARMA

As the univariate benchmark, the ARMA model was unsurprisingly the worst-performer in test periods 1 and 3, performing better than only the tree-based models in test period 2. In tranquil economic periods, the model performs fairly well. In test period 3 before the first quarter of 2020, when GDP growth rates hewed very close to their long-term averages with minimal variation, the ARMA model was actually the best performer. However, its weakness was apparent during periods with large shocks. It predicted large declines in the third quarter of 2020, belatedly reacting to the strong decline of the second quarter of 2020 even though the quarter saw the largest growth rate recorded in the series. The approach's usefulness as a standalone methodology is limited in nowcasting for all but the most predictable series, as evidenced by its second-worst showing in figure 4.

### Bayesian VAR

The Bayesian VAR model was highly performant in the analysis, ranking as the best performer for test period 1 and second-best performer for test periods 2 and 3 in terms of both MAE and RMSE. It ranked second in figure 4, though its score was closer to third place MIDAS than to the first place LSTLM. Its ability to pick up both 2020 Q2's large decline and 2020 Q3's large growth was rivalled only by the LSTM. The methodology produced, however, the third-most volatile predictions in test period 1 and the most volatile predictions in test periods 2 and 3. High volatility in a nowcast can be an issue, as large revisions in predictions over time can make it difficult to communicate and rely on forecasts.

### Decision Tree

The decision tree did not perform particularly well in the analysis, especially at earlier data vintages. In test period 1, the tree learned most of its information from the last month of each quarter, so it was only able to start producing estimates significantly different from the mean at the one month after vintage. Its predictions were quite accurate at the two months after vintage, but this pattern did not carry over to the second and third test periods, where its predictions were not very accurate even at the two months after vintage. Importantly, it was unable to produce predictions significantly higher or lower than those it had previously seen in training, unable to produce estimates of large contraction in 2020 Q2 despite the strong negative signals in the data. All of these shortcomings led to an analysis-worst score in figure 4.

### DFM

The DFM performed exceptionally well in the first test period, bested only by the Bayesian VAR. This high performance did not carry over to the second two test periods, however, where it ranked sixth for both. It displayed a positive bias in both the second and third test periods, while at points severely overestimating the degree of contraction in 2020 Q2, and underestimating growth in 2020 Q3. In testing, the DFM performed better in the second two test periods provided with only the variables used for test period 1, indicating the methodology may be better suited when few explanatory variables are at hand. Overall, it ranked fourth in figure 4, exhibiting similar performance to ridge regression.

### Gradient Boosted Trees

Among the poor-performing tree-based methodologies, gradient boost was a middling performer, ranking between random forest and decision tree in every test period. Like the other tree-based methods, gradient boost seemed to learn most of its information from the latest available data, making its two months after vintage predictions quite accurate, but its earlier predictions hewed very close to the mean. Like the decision tree, it was unable to predict anything outside scales it had encountered before, only able to predict small contractions in 2020 Q2. One approach tested to mitigate this behavior which particularly benefited gradient boost was training a different model for each data vintage. That is, training five separate models, each only with information that would be available at that vintage. This forces the model to learn more information on data available earlier. Full details and specifications can be found in Hopp (2022).

### LSTM

The LSTM performed well in all three test periods, ranking third-best in the first and best in the second two. Performance was especially strong in the COVID crisis. It was rivalled by perhaps only MIDAS in terms of combining accurate predictions with low volatility in revisions, which together led to an analysis-best aggregate performance in figure 4, by a wide margin.

### MF-VAR

The MF-VAR model was accurate in the two months after vintage, but struggled in earlier data vintages and during the COVID crisis, producing some outlier predictions in 2020 Q2 and 2021 Q1. It was the second-most volatile methodology after OLS in the test period 1 and Bayesian VAR in the second and third test periods. Overall, it performed better than OLS, ARMA, and the tree-based methodologies in figure 4, ranking seventh.

### MIDAS

MIDAS was a well-performing methodology, ranking behind only the LSTM and Bayesian VAR in terms of overall performance in figure 4. It had trouble predicting extreme values during both the financial crisis and COVID crisis, but was particularly non-volatile in its revisions, ranking fourth-least volatile in the first test period and third-least in the second two.

### MLP

The MLP was exceptionally poor in test period 1, ranking as the worst-performing model in terms of MAE for the period and besting only ARMA in terms of RMSE. Performance in the second two test periods was much improved, however. It was the fifth-best performing model in test period 2 and fourth-best in period 3. It accurately predicted 2020 Q2's decline, but struggled, as many methodologies, in predicting 2020 Q3's rebound. Overall, it ranked sixth in figure 4.

### OLS

Given the high number of variables plus lags included in the modelling, it is not surprising to see OLS as a relatively weak and volatile performer in the analysis. The exception is test period 3, where it did quite well in prediction during 2020 Q2 and Q3 and was actually the third-best performing model, this despite an outlier prediction in 2020 Q1. Despite strong performance in the final test period, it beat out only ARMA and the decision tree in terms of overall performance in figure 4.

### Random Forest

Random forest suffered from the same issues as gradient boosted trees, having difficulty predicting anything other than the mean until the two months ahead vintage and unable to predict extreme values in test period 3. It was the best-performing tree-based methodology, slightly edging out gradient boost in figure 4.

**Ridge Regression**

Ridge regression produced middling results in the analysis, ranking sixth in test period 1, third in test period 2, and eighth in test period 3. Volatility was significantly mitigated compared with OLS, as expected due to the regularization parameter, leading to an overall ranking of fifth in figure 4. Though its aggregate score was closer to second-ranked Bayesian VAR than to seventh-ranked MF-VAR.

# 6. Conclusion

Never before have nowcasting practitioners had so many options when it comes to selecting which methodology to use in their applications. This paper has attempted to ease that selection process by providing comparative results in nowcasting US GDP growth across three different volatile periods in US economic history. The two best-performing methodologies across all three periods were found to be the LSTM and Bayesian VAR, the latter coming with the caveat of having the highest revisions upon receiving new data in the analysis.

A primary aim of this paper was not only comparing methodologies on a benchmark dataset, but enabling others to use and explore those methodologies themselves. To that end, the exact code used to produce these results for each methodology is published on GitHub at Hopp (2022). This includes not only the function calls together with hyperparameters, but the code for getting the data and transforming it, the code necessary for generating and testing on artificial lags, and the code for generating predictions with new data. Once one's data is in a standardized format, with a date column, observations in rows, and series in separate columns, they can follow the self-contained boilerplate code in each methodology's Jupyter Notebook to adapt it to their own application or use as a springboard for their own research.

Beyond predictive performance, other factors such as computation time, time frequencies handled, and implementations available need to be take into account when selecting a nowcasting methodology. More information on these specific characteristics in the implementations used for this analysis are available at Hopp (2022). With more tools, transparency, and context, existing practitioners may be able to try different nowcasting approaches to validate their models and new ones may be able to enter the field with novel applications.